\documentclass[sn-mathphys,Numbered]{sn-jnl}

\usepackage{graphicx}%
\usepackage{multirow}%
\usepackage{amsmath,amssymb,amsfonts}%
\usepackage{amsthm}%
\usepackage{mathrsfs}%
\usepackage[title]{appendix}%
\usepackage{xcolor}%
\usepackage{textcomp}%
\usepackage{manyfoot}%
\usepackage{booktabs}%
\usepackage{algorithm}%
\usepackage{algorithmicx}%
\usepackage{algpseudocode}%
\usepackage{listings}%
\usepackage{makecell}

\theoremstyle{thmstyleone}%
\theoremstyle{thmstyletwo}%

\theoremstyle{thmstylethree}%

\begin{document}

\title[Article Title]{Item Region-based Style Classification Network (IRSN): A Fashion Style Classifier Based on Domain Knowledge of Fashion Experts}


\author[1]{\fnm{Jinyoung} \sur{Choi}}\email{jychoi@handong.ac.kr}

\author[1]{\fnm{Youngchae} \sur{Kwon}}\email{youngchaekwon@handong.ac.kr}
\equalcont{These authors contributed equally to this work.}

\author*[1]{\fnm{Injung} \sur{Kim}}\email{ijkim@handong.edu}
\equalcont{These authors contributed equally to this work.}

\affil*[1]{\orgdiv{School of CSEE}, \orgname{Handong Global University}, \orgaddress{\street{558 Handong-ro Buk-gu}, \city{Pohang}, \postcode{37554}, \state{Gyeongbuk}, \country{Republic of Korea}}}


\abstract{Fashion style classification is a challenging task because of the large visual variation within the same style and the existence of visually similar styles. 
    Styles are expressed not only by the global appearance, but also by the attributes of individual items and their combinations.
    In this study, we propose an item region-based fashion style classification network (IRSN) to effectively classify fashion styles by analyzing item-specific features and their combinations in addition to global features.
    IRSN extracts features of each item region using item region pooling (IRP), analyzes them separately, and combines them using gated feature fusion (GFF).
    In addition, we improve the feature extractor by applying a dual-backbone architecture that combines a domain-specific feature extractor and a general feature extractor pre-trained with a large-scale image-text dataset.
    In experiments, applying IRSN to six widely-used backbones, including EfficientNet, ConvNeXt, and Swin Transformer, improved style classification accuracy by an average of 6.9\% and a maximum of 14.5\% on the FashionStyle14 dataset and by an average of 7.6\% and a maximum of 15.1\% on the ShowniqV3 dataset. Visualization analysis also supports that the IRSN models are better than the baseline models at capturing differences between similar style classes.}


\maketitle

\section{Introduction}\label{sec1}

Since the advent of deep learning, image processing technology has developed rapidly, and there have been increasing attempts to apply it to the fashion field.
Recently, the online fashion markets, such as apparel shopping apps and web services, have been growing rapidly \cite{ecommerce}.
In general, in online fashion shopping malls, sellers display images of people wearing fashion items such as clothes, hats, shoes, and accessories, so that users can select the items.
Since different users have different style preferences, recognizing the style of fashion images and recommending items that match the user's taste can greatly improve user satisfaction.
To this end, a style classifier that can effectively classify the style of fashion images is essential.

\begin{figure}[h]%
\centering
\includegraphics[width=0.9\textwidth]{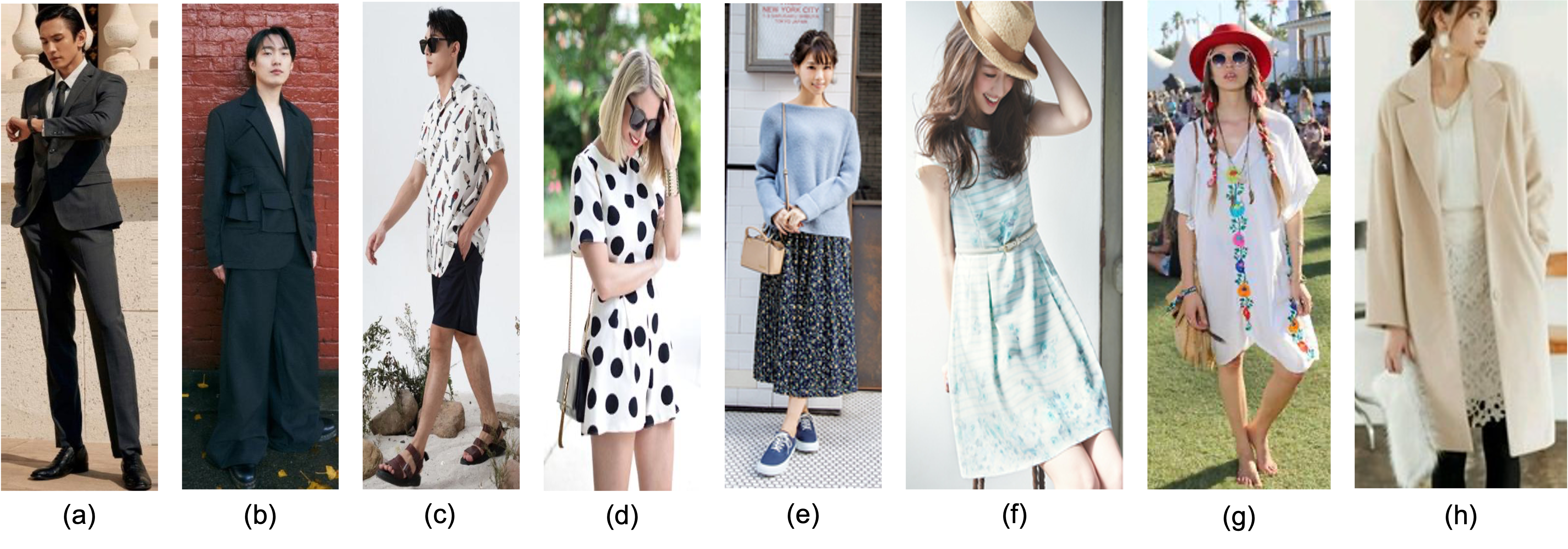}
\caption{Examples of fashion styles. (a) classic (gentleman), (b) minimalist  (c) vacation, (d) retro, (e) girly, (f) feminine, (g) ethnic, and (h) feminine (with different appearance from (f)).}\label{fig1}
\end{figure}

Fashion can be broadly understood as shifting styles of dress, which are specific combinations of silhouettes, textiles, colors, details, and fabrications that are embraced by groups of people at a particular time and place \cite{kennedy2013fashion}.
There are eight traditional fashion styles: classic, romantic, elegance, sophisticated, modern, mannish, active, and ethnic \cite{funda_fashion}. Over time, new styles have been added, such as vacation, girly, and retro, and there are a large number of style types in use. Fig.~\ref{fig1} displays examples of fashion images and their style labels and Table~\ref{tab1} lists the fashion styles in the FashionStyle14 \cite{TakagiICCVW2017} and ShowniqV3 datasets. Fashion experts distinguish between them by the types of items worn, colors, embellishments, patterns, and materials \cite{kennedy2013fashion}. For example, the classic style is represented by items such as tailored suits, Chanel suits, and cardigan sweaters, materials such as velvet and tweed, and timeless colors such as neutrals and browns, while the ethnic style is represented by combinations of multiple colors, folk-inspired materials and patterns, and colorful items \cite{funda_fashion}.

\begin{table}[t!]
\caption{Fashion styles in the FashionStyle14 and ShowniqV3 datasets}\label{tab1}
\begin{tabular*}{\textwidth}{@{\extracolsep\fill}ccl}
\toprule%
Dataset & Style & Description \\
\midrule  
\multirow{3}{*}{FashionStyle14} & Conservative & \makecell[l]{Avoid revealing clothing. Calm colors such as black,\\ white, gray, brown, and navy.}\\
\cmidrule{2-3}
& Fairy	& Oversized tops and pastel colors.\\
\cmidrule{2-3}
& Feminine & \makecell[l]{Emphasize the waistline and leg line. Decorated with \\ruffle and lace. Chiffon and silk materials.}\\ 
\cmidrule{2-3}
& Gal & \makecell[l]{Items such as short skirts and high shoes; makeup that\\ emphasizes brown hair color, tan skin, and red cheeks.}\\
\cmidrule{2-3}
& Kireime-casual & Semi-formal and businesslike items; subdued colors.\\
\cmidrule{2-3}
& Lolita & \makecell[l]{Colors such as black, ivory, pink, and blue; accessories \\like a headbow and wristcuff; and a voluminous skirt.}\\
\cmidrule{2-3}
& Mode & \makecell[l]{Distinctive silhouette mimicking high fashion; black or\\ solid color.}\\
\cmidrule{2-3}
& Natural & \makecell[l]{Materials such as cotton or linen; calm tones such as\\ white, navy, or beige.}\\
\cmidrule{2-3}
& Rock & \makecell[l]{Embellishments such as studs, sequins, faux leather;\\slim fit or straight cuts; dark colors such as black, gray.}\\
\cmidrule{2-3}
& Dressy & \makecell[l]{Velvet, satin and other materials; long dresses that\\ emphasize the waistline and shoulders.}\\
\midrule
\multirow{3}{*}{common} & Ethnic (Bohemian)	& \makecell[l]{Harmony of multiple colors, folk-style materials or\\ patterns, and items with colorful patterns.}\\
\cmidrule{2-3}
& Girly (Girlish) & \makecell[l]{Ruffles, lace, gather decorations; check or flower \\patterns; pink, white, or pastel colors.}\\
\cmidrule{2-3}
& Retro	& \makecell[l]{Dot, stripe and check patterns; bootcuts and low-\\waisted bottoms; accessories like hairband and scarf.}\\
\cmidrule{2-3}
& Street & \makecell[l]{Loose and wide silhouette; Items such as beanies and \\bucket hats.}\\
\midrule
\multirow{3}{*}{ShowniqV3} & Classic (Gentleman) & \makecell[l]{Items such as tailored suits, cardigans, and sweaters;\\ velvet and tweed materials; neutral colors and browns.}\\
\cmidrule{2-3}
& Minimalist (Modern) & \makecell[l]{Monotone or neutral color; Simple design with no \\zippers, pockets or buttons.}\\
\cmidrule{2-3}
& Vacation & \makecell[l]{Tropical patterns; Rattan or mesh materials; Items like\\ sunglasses and sandals.}\\
\cmidrule{2-3}
& Sporty & Leggings, track pants, and sleeveless items.\\
\cmidrule{2-3}
& Tech & \makecell[l]{Decorations such as pockets, buttons, zippers, velcro;\\ Gore-Tex or Cordura materials; Items like cargo pants.}\\
\cmidrule{2-3}
& Punk & \makecell[l]{Metal chain decorations; leather or mesh materials and\\ flashy prints; Items such as work boots, long boots, and\\ leather jackets.}\\
\cmidrule{2-3}
& Basic	& \makecell[l]{Standard fit; Items such as sneakers, jeans, and shirts.}\\
\botrule
\end{tabular*}
\label{tab:styles}
\end{table}

Deep learning has shown impressive performance in image processing. Convolutional neural networks (CNNs) have long been widely used in various computer vision tasks, such as image classification, object detection, and segmentation \cite{resnet, googlenet, mobilenetv2, efficientnet, fastrcnn,yolo,ssd,hrnet,convnext,convnext2}. In recent years, image processing models based on Transformers \cite{transformer}, which have achieved great success in natural language processing, are also dominant in computer vision \cite{vit,swin,swin2,ccnet}.

However, it is difficult to achieve good performance by applying a neural network designed for general object recognition to fashion style classification \cite{style_difficult}.
Fashion style recognition is a fine-grained classification task that requires delicately distinguishing subtle differences between styles from fashion images of diverse shapes.
Since each style can be represented by items with various shapes and colors, even fashion images of the same style can have significantly different visual appearances, while different styles may have similar appearances. 
For example, in Fig. \ref{fig1}, both (f) and (g) show a woman wearing a hat and dress and have similar global shapes. (f) is classified as feminine style because the hat, dress, and belt that accentuates the waist create a feminine mood, while (g) is ethnic style because the intensely colored hat and dress pattern create a folkloric atmosphere. On the other hand, (h) belongs to the same feminine style as (f), but has a significantly different visual appearance from (f). 
In order to accurately capture the characteristics of each style and the differences between them, it is necessary to take into account not only global features such as silhouettes, but also the attributes of individual items and the harmony of their combination.

Consequently, in order to effectively recognize fashion styles using deep learning, we need a network architecture that can effectively analyze not only global features but also item-level features and also need data to train the model. However, there is no dataset that provides both style labels and item-level attribute labels that can be directly used to learn correlations between styles and item attributes. For example, the women's fashion style dataset FashionStyle14 \cite{TakagiICCVW2017} provides style labels but not attribute labels for each item. Another public dataset, DeepFashion \cite{deepfashion}, provides item-level attribute labels but not fashion style labels.

In this paper, we propose an item region-based style classification network (IRSN) to effectively recognize fashion styles by combining global features with item-level regional features.
As illustrated in Fig.~\ref{fig2}, IRSN applies a pre-trained item segmentation model and the item region pooling (IRP) which is proposed in this paper to separate the features of head, top, bottom, and shoes item regions from the global feature map extracted by the backbone network, and analyzes each of them using separate item encoders. Then, it combines the global and item-specific features through gated feature fusion (GFF) to recognize the style of the input image. GFF multiplies the features extracted from multiple items by their weights and then combines them, where the weight of each feature is dynamically computed according to the input data. Therefore, high weights are assigned to item features that are important for style recognition, while low weights are assigned to features of unimportant or undetected items.
Additionally, we applied a dual-backbone consisting of a domain-specific feature extractor trained with fashion style data and a general feature extractor trained with a large-scale general image-text dataset to improve feature extraction performance and compensate for the lack of style-labeled data.

Applying IRSN to six backbone networks widely used in computer vision, including EfficientNet, ConvNeXt, and Swin Transformer, improved accuracy on the FashionStyle14 women's style data by an average of 6.9\% and a maximum of 14.5\% compared to the corresponding baseline models. On the ShowniqV3 dataset men's style data, the improvement was an average of 7.6\% and a maximum of 15.1\%. These results clearly show that the proposed structure of IRSN, which separates item-specific features by IRP, analyzes them separately, and then integrates them by GFF, is effective for fashion style classification.

\section{Related Work}\label{sec2}

\subsection{Deep learning models for image processing}\label{subsec2_1}

Since AlexNet \cite{alexnet} won the ImageNet Challenge, CNNs have been widely used in image processing. 
VGG, Inception, and ResNet \cite{vgg16, googlenet, resnet} significantly improved the performance of CNNs by applying more layers, parallel structures, auxiliary classifiers, skip connections, etc. 
There has also been a lot of research on reducing the computation and memory usage of CNNs. MobileNet replaces full-convolution with separable convolution that requires less computation and parameters \cite{mobilenet}, and MobileNetV2 further improved efficiency by applying inverted residual block and linear bottleneck \cite{mobilenetv2}. EﬀicientNet exhibited improved performance with fewer parameters than existing models by maintaining the balance between scaling in network depth, width, and resolution. 
In recent years, researchers have been actively applying Transformer networks \cite{transformer}, which have shown excellent performance in the field of natural language processing, to computer vision.
Vision Transformer (ViT) outperformed CNNs on the ImageNet dataset for the first time for a Transformer-based model \cite{vit}. Swin Transformer extends ViT to a hierarchical structure by applying shifted windows \cite{swin, swin2}. 
Composed of a multi-scale architecture, Swin Transformer was designed as a backbone network for various image processing tasks such as object detection and segmentation as well as classification. Transformers are also widely used for unsupervised and self-supervised pre-training of image processing models \cite{simmim, mae}.

There has also been a lot of research on designing efficient and effective architectures by combining the advantages of CNNs and Transformers. Dai et al. \cite{coatnet} proposed CoAtNet, which improves performance by combining depthwise convolution and self-attention. Park and Kim analyzed the characteristics of convolution and self-attention and proposed an AlterNet that maximizes the synergy between the two operations by arranging convolution and self-attention alternately \cite{howvit}.
Liu et al. proposed ConvNeXt, a CNN architecture modernized by applying the improvement factors of Transformers to ResNet50 \cite{convnext}. Later, Touvron et al. proposed data-eﬀicient image Transformer (DEIT) \cite{deit} that applied distillation specialized for Transformers, and Yang et al. proposed focal self-attention to efficiently and effectively capture short- and long-range visual dependencies \cite{focal}.

There are lots of studies on building lightweight Transformers for limited environments, such as mobile phone. Ho et al. reduced the computational requirements of Transformers using axial attention \cite{axialatt}, and Huang et al. proposed criss-cross attention to learn contextual information efficiently and effectively \cite{crisscross}. Wan et al. proposed SeaFormer, which can trade-off between accuracy and latency using a squeeze-enhanced axial Transformer \cite{seaformer}.

\subsection{Fashion Style Classification}\label{subsec2_2}

There are few prior studies on fashion style classification. In \cite{fashiongraph}, Kim et al. proposed a multi-label fashion style classification method using graph convolution network (GCN) and transfer learning. They improve feature extraction performance by transferring knowledge from the pre-trained ResNet \cite{resnet} for the ImageNet dataset. They apply GCN to learn and utilize the correlation between styles.
Chen et al. proposed dual attention that combines criss-cross attention \cite{crisscross} and spatial attention \cite{spatial} to improve fashion style classification performance \cite{dualattention}. The criss-cross attention module \cite{crisscross} mainly reflects the relationship between adjacent pixels, and the spatial attention \cite{spatial} module determines the attention weights on the spatial dimension.

\subsection{Pre-trained Vision-Language models}\label{subsec2_3}

Transfer learning, which transfers the knowledge of a model pre-trained on a large-scale dataset and then fine-tunes to the target data, is a popular approach to improve the performance of feature extractors \cite{pretrained},
Many previous studies pre-trained neural networks on object classification datasets such as ImageNet \cite{preimagenet1, preimagenet2, preimagenet3}.
Recently, many researchers proposed pre-training methods to acquire knowledge from image-text pair datasets using vision-language models \cite{visual_ngram, weaklysup_bow, convirt, clip}. Since text descriptions provide richer information than class labels, pre-training with vision-language models can not only acquire more knowledge than pre-training with classification models. In addition, vision-language models can be used for multimodal applications such as text-to-image transfer, as they learn a common embedding space between texts and images.

Li et al. proposed a visual n-gram model that learns knowledge from images on the Web by predicting the n-grams in user comments associated with the images \cite{visual_ngram}.
Joulin et al. convert text labels into bag-of-words (BoW) and pre-train the model to predict them \cite{weaklysup_bow}.
Zhang et al. proposed ConVIRT, a model that learns to match the embeddings of medical images and their text descriptions through contrastive learning \cite{convirt}.
Recently, Radford et al. proposed CLIP (Contrastive Language-Image Pre-Training), a network that takes a batch of (image, text) pairs as input and trains an image encoder and a text encoder to predict the correct pairings between image and text within the batch \cite{clip}. CLIP exhibited a significant improvement in training efficiency over previous methods and is effective for pre-training on large-scale data. CLIP exhibits excellent zero-shot prediction performance. In a following study \cite{clipseg}, Lüddecke et al. proposed CLIPSeg, a zero-shot segmentation model using CLIP.

\begin{figure}[t!]%
\centering
\includegraphics[width=1.0\linewidth]{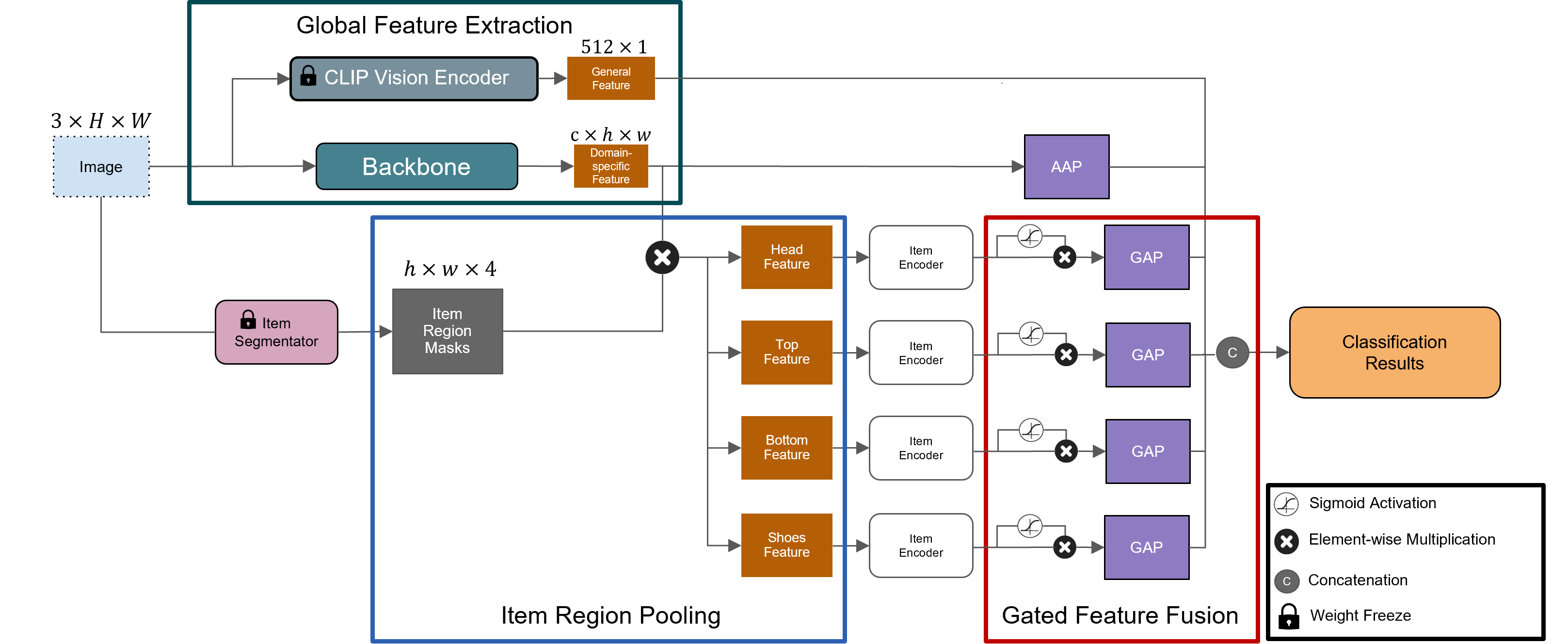}
\caption{The structure of IRSN. $(H, W)$ and $(h, w)$ denote the size of the input image and the feature map, respectively.}\label{fig2}
\end{figure}

\section{Item Region-based Style Classification Network}\label{sec3}

The overall structure of the proposed model IRSN is shown in Fig.~\ref{fig2}. IRSN first extracts global features using a dual backbone consisting of a domain-specific feature extractor and a general feature extractor. Then, it segments the head, top, bottom, and shoes regions using a pre-trained item segmentation model and extracts the feature map of each item region from the global feature map through item region pooling (IRP). Item encoders receive item-specific feature maps as input and output higher-level item feature maps and spatial attention maps. IRSN combines item-specific features and global features through gated feature fusion (GFF). Finally, the classification head predicts the fashion style of the input image from the combined feature vector.

\subsection{Global Feature Extraction}\label{subsec3_1}

The global feature extractor receives images of people wearing fashion items and extracts global features that reflect the overall shape, color, and other global attributes. IRSN's global feature extractor consists of a dual-backbone architecture that combines a domain-specific feature extractor (DFE) and a general feature extractor (GFE). The former is trained on fashion image data to extract feature maps specialized for fashion image analysis, while the latter is trained on a large-scale general image-text dataset to extract feature vectors useful for various downstream tasks.

The domain-specific feature extractor $DFE(\cdot): \mathbb{R}^{3 \times H \times W} \mapsto \mathbb{R}^{c \times h \times w}$ extracts a feature map $d_x \in \mathbb{R}^{c \times h \times w}$ from a color image $x \in \mathbb{R}^{3 \times H \times W}$, as shown in Eq.~(\ref{eq1}), where $c$, $h$, and $w$ depend on the type of the backbone.
\begin{equation}
    d_x = DFE(x)
\label{eq1}
\end{equation}

In this study, we applied six networks widely used in image processing as backbones: ResNet50 \cite{resnet}, ConvNeXt-Tiny \cite{convnext}, ConvNeXt-Base \cite{convnext}, EfficientNet-B3 \cite{efficientnet}, MobileNetV2 \cite{mobilenetv2}, and Swin Transformer-Base \cite{swin}. To improve feature extraction performance and robustness, we transferred parameters pre-trained on ImageNet and then fine-tuned them to the fashion image data, integrated with other modules.

In addition, we combined a general feature extractor to further improve feature extraction performance by using knowledge learned from a large-scale general image-text dataset. As shown in Eq.~(\ref{eq2}), the general feature extractor $GFE(\cdot): \mathbb{R}^{3 \times H \times W} \mapsto \mathbb{R}^D$ extracts a general feature vector $g_x$ from the input image.
\begin{equation}
    g_x = GFE(x)
\label{eq2}
\end{equation}

In this study, we used the vision encoder of CLIP (Contrastive Language-Image Pre-Training) \cite{clip}, a large-scale vision-language model trained with 400M (image, text) pairs, as a general feature extractor.
Because CLIP learns the correlations between images and text, it learns more diverse and generalizable knowledge than models trained only on images and class labels, making it a useful complement to domain-specific feature extractors trained on fashion images. Since CLIP learns a common embedding space for both images and text, the CLIP vision encoder outputs a linear feature vector compatible with text embedding ($D=512$) instead of a 3D feature map. During training, we fixed the parameters of the CLIP vision encoder to preserve its general knowledge.

\begin{figure}[h] 
\centering
\includegraphics[width=0.9\textwidth]{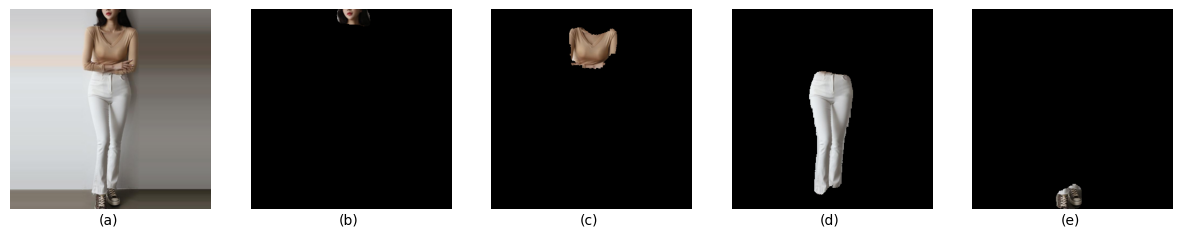}
\caption{An example of item segmentation result. (a) input image, (b)-(e) the masks of head, top, bottom, and shoes regions element-wise multiplied by the input image, respectively.}\label{fig3}
\end{figure}

\subsection{Item Feature Extraction}\label{subsec3_2}

Since fashion style is influenced by the attributes of individual items, IRSN analyzes the features of each item separately. IRSN first generates binary masks of four regions, head, top, bottom, and shoes, from the input image using a pre-trained item segmenter, as shown in Eq.~(\ref{eq3}). In this study, we used CLIPSeg \cite{clipseg} because of its excellent zero-shot performance. Fig.~\ref{fig3} shows examples of head, top, bottom, and shoes region masks ((b)-(e)) extracted from input image (a).
\begin{equation}
m_i = CLIPSeg(x, prompt_i),
\label{eq3}
\end{equation}
where $m_i \in \{0, 1\}^{H \times W}$ is the mask of an item $i \in \{head, top, bottom, shoes \}$. We empirically chose ( ``head", ``casual top cloth", ``pants, skirt cloth", ``shoes") as the prompts to retrieve the item regions. 

\begin{figure}[h] 
\centering
\includegraphics[width=0.9\columnwidth]{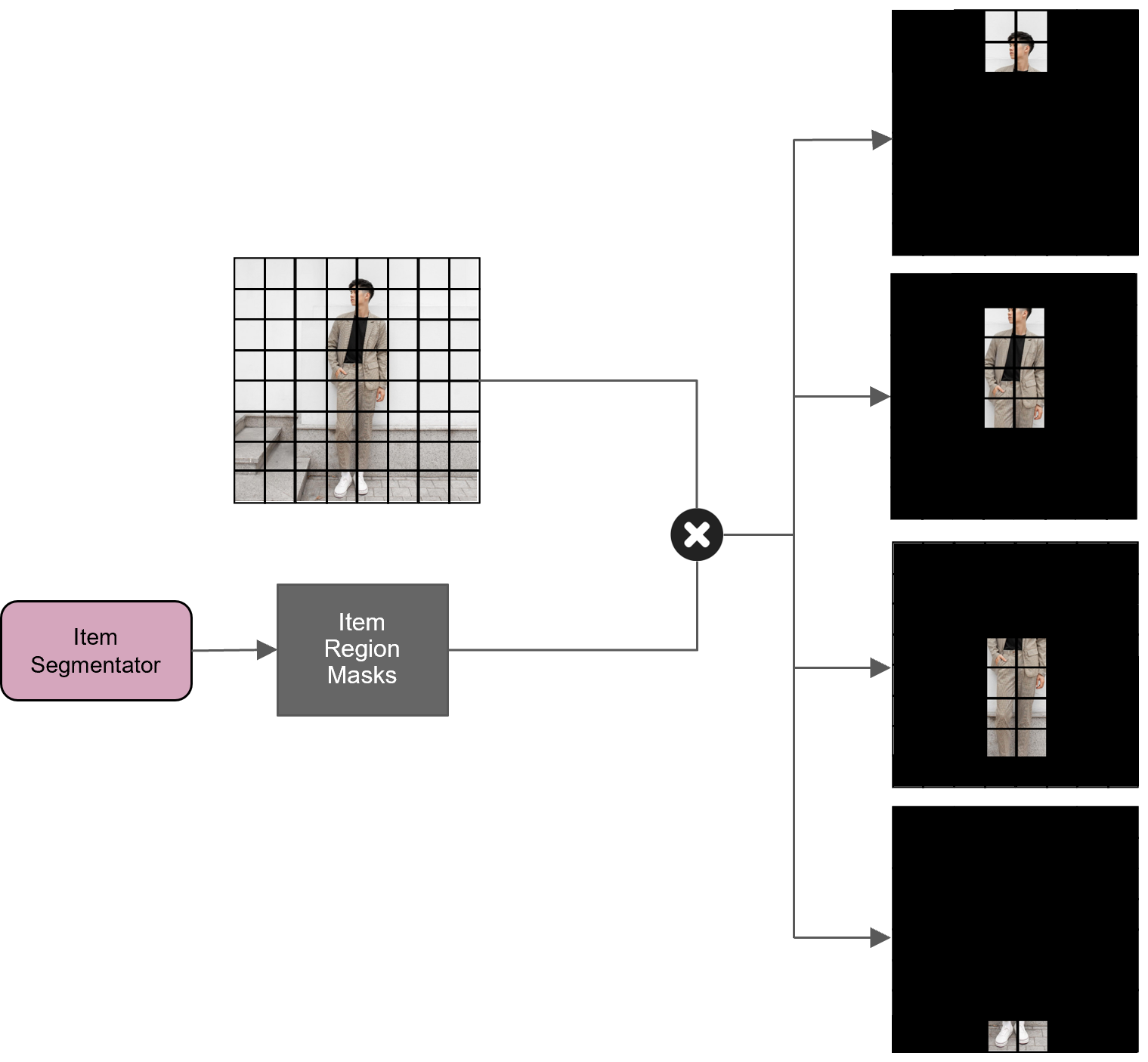}
\caption{Item Region Pooling. The binary mask generated by the item region segmentator is resized to the size of the feature map extracted by the backbone for domain-specific feature extraction and then multiplied to pool only the item area for each region.}\label{fig4}
\end{figure}

Then, we separated the feature map of each item region from the global feature using item region pooling (IRP). As illustrated in Fig.~\ref{fig4}, IRP retrieves the feature of each item by multiplying the downsampled item region map with the domain-specific global feature $d_i$.
The operation of IRP is presented in Eq.~(\ref{eq4}).
\begin{equation}
    f_i(x) =  m_i \downarrow_{h \times w} \otimes d_x,
\label{eq4}
\end{equation}
where $\downarrow_{h \times w}$ denotes downsampling to $h \times w$ size and $\otimes$ denotes element-wise multiplication that broadcasts the smaller feature map across the larger feature map.
We applied the item encoder to further analyze the features of each item, as shown in Fig.~\ref{fig2}.
\begin{equation}
    [h_i(x), a_i(x)] = ItemEnc(f_i(x))
\label{eq5}
\end{equation}
Each item encoder takes $f_i$ as input and produces a higher-level feature map $h_i \in \mathbb{R}^{c \times h \times w}$ as shown in Eq.~(\ref{eq5}). The item encoder also produces an importance map $a_i(x) \in [0,1]^{c \times h \times w}$ for the corresponding item, which will be used to combine item features as described in Subsection~\ref{subsec3_3}. In this study, we implemented the item encoders with Transformer blocks \cite{transformer}. To allow greater importance weights to certain items than others, we applied the Sigmoid activation to $a_i(x)$ instead of the Softmax activation. If an item is absent in the input image or undetected by the item segmentator, we set $a_i(x)$ to zero for all coordinates.

\begin{figure}[h] 
\centering
\includegraphics[width=0.5\columnwidth]{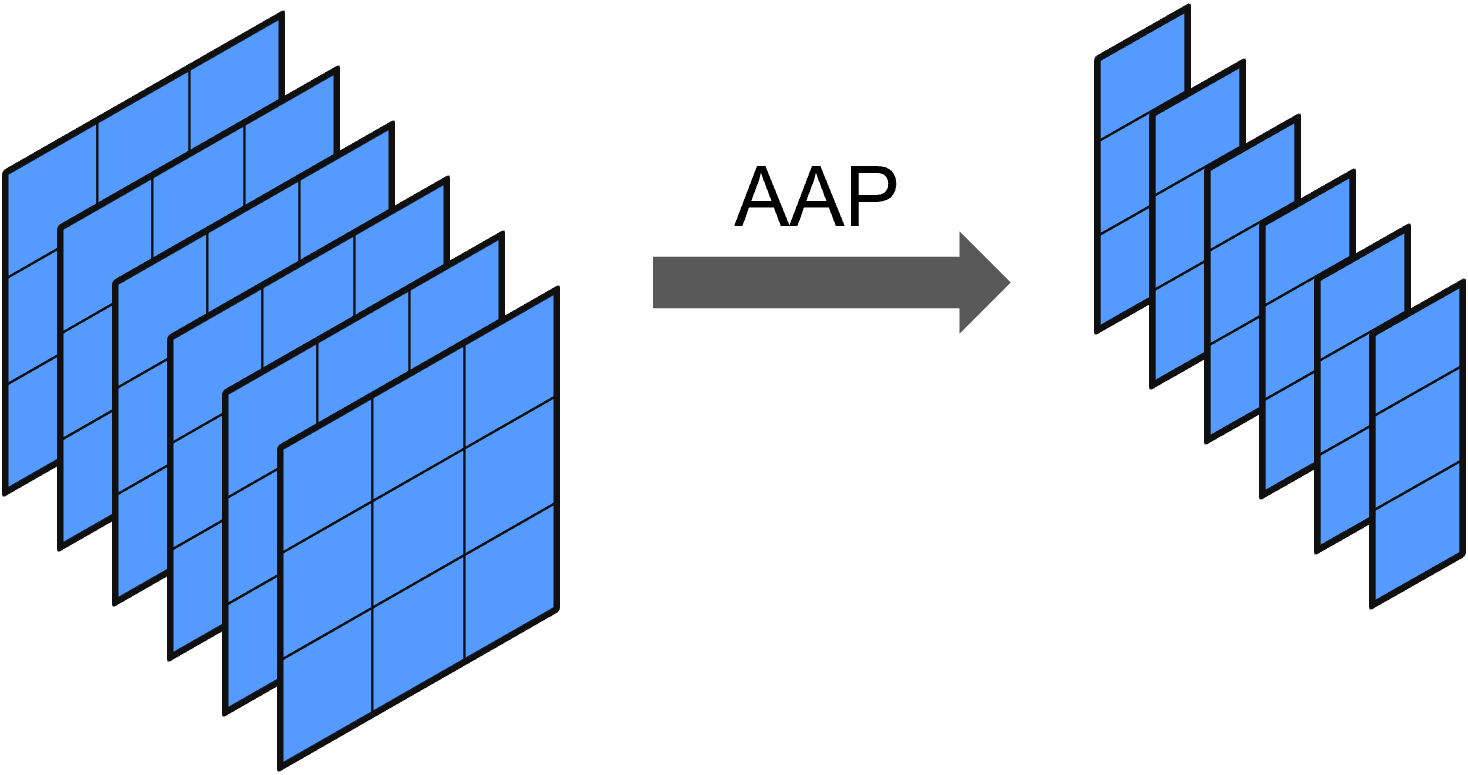} 
\caption{Adaptive average pooling from $c \times 3 \times 3$ to $c \times 3 \times 1$}\label{fig5}
\end{figure}

\subsection{Gated Feature Fusion and Style Classification}\label{subsec3_3}

To predict the style of the input image, IRSN combines three types of features: domain-specific global features, general global features, and item-specific features. In order to integrate these features with different shapes, we first reshape the item feature maps into linear vectors and then concatenate them.
Since concatenating multiple features results in excessively high dimensions, we reduced the dimensionality of the features by pooling operations before combining them.

In general, image recognition models apply global average pooling (GAP) to a high-level layer. While GAP reduces overfitting and induces a strong correlation between the channels of the feature map and the target classes, it has the disadvantage of losing positional information.
Positional information is important for fashion style classification because, in fashion images, the position of fashion items is strongly biased towards certain regions. Therefore, we used adaptive average pooling (AAP) instead of GAP to preserve the positional information. Since fashion images usually have a vertically long shape, we applied $5\times1$ or $5\times3$ as the output resolution of AAP.

IRSN extracts feature maps from multiple item regions, as shown in Fig.~\ref{fig2}. However, the importance of each item region in style classification varies depending on the input image and its style. In addition, certain items may be absent in the input image or undetected by the segmentator. To deal with these issues, we applied gated feature fusion (GFF) to integrate item features reflecting the presence and importance of each item. GFF scales item feature maps $h_i(x)$ by multiplying them with the corresponding importance maps $a_i(x)$, which were estimated by the item encoders, reduces the dimension through GAP, and then concatenates the reduced features.
\begin{equation}
    GFF_{i \in items}(h_i(x), a_i(x)) = \\        
    Concat_{i \in items}(GAP(h_i(x) \otimes a_i(x))),
    \label{eq6}
\end{equation}
where $items=\{head, top, bottom, shoes\}$. We applied GAP instead of AAP because each item region is a specific part of the entire image. The features of items that are absent or undetected are ignored because IRSN assigns them zero importance.

Since the general global features extracted by the CLIP vision encoder already have a moderate size, we did not apply any pool operation. After scaling to reflect the importance of each feature and reducing the dimensionality, we combined them to predict the style. We implemented the classification head with a multi-layer perceptron (MLP) with $K$ output nodes, where $K$ is the number of style classes. The style $s(x)$ of an input image x is determined by Eq.~(\ref{eq7}).
\begin{align} \label{eq7}
    \begin{split}
        y(x) &= MLP(Concat(AAP(d_x), g_x, GFF_{i \in items}(h_i(x), a_i(x)))) \\    
        s(x) &= argmax_{k \in [1,K]} y(x)_k         
    \end{split}
\end{align}

We used the cross entropy loss to train the model. The detailed training procedure is described in Subsection~\ref{subsec4_2}.

\section{Experiments}\label{sec4}

\subsection{Dataset and evaluation metric}\label{subsec4_1}

We used the FashionStyle14 dataset \cite{TakagiICCVW2017} and ShowniqV3 dataset in experiments. FashionStyle14 is a Japanese women's style classification dataset consisting of 14 style classes: conservative, dressy, ethnic, fairy, feminine, gal, girlish, kireime-casual, lolita, mode, natural, retro, rock, and street. We used 12,711 images from FashionStyle14 \cite{TakagiICCVW2017} for training and evaluation, excluding blank or corrupted images. ShowniqV3 is an in-house style image dataset collected by a fashion platform company, DeepFashion. ShowniqV3 consists of 2,796 men's style images categorized into 10 style classes: gentleman, bohemian, minimalist, vacation, sporty, techwear, retro, street, punk, and basic. Each sample in both datasets is assigned only one style label.

\subsection{Training details}\label{subsec4_2}

We used the six backbone networks listed in Subsection~\ref{subsec3_1} for the domain-specific feature extractor. For each backbone, we transferred the parameters pre-trained on ImageNet \cite{imagenet}, integrated with other modules and then fine-tuned them on the target dataset. For the general feature extractor, we applied the CLIP vision encoder \cite{clip} pre-trained on 400M (image, text) pairs and froze its parameters to preserve the general knowledge learned from the large-scale dataset.
We used the cross entropy loss for training and applied label smoothing \cite{labelsmoothing} to prevent excessive competition between visually similar style classes.
We optimized the parameters using the stochastic gradient descent (SGD) optimizer with a batch size of 16, a learning rate of 1e-5, and a momentum of 0.9.

We used the pre-trained CLIPSeg \cite{clipseg} as the item region segmentator without fine-tuning. CLIPSeg takes an image and a text prompt as input and outputs a segmentation map. The prompts used to extract each item region are described in Subsection~\ref{subsec3_2}.

\subsection{Style classification accuracy}\label{subsec4_3}

\begin{table}[h]
\caption{The style classification accuracy of six IRSN models and their baselines on the FashionStyle14 \cite{TakagiICCVW2017} and ShowniqV3 datasets. IRSN exhibited improved accuracy for all six backbones. The average improvement for the two data sets was 6.9\% and 7.6\%, respectively. In particular, the `EfficientNet-B3 + IRSN' model outperformed the baseline model by 14.5\% on FashionStyle14 \cite{TakagiICCVW2017} and by 15.1\% on ShowniqV3.}\label{tab2}
\begin{tabular*}{\textwidth}{@{\extracolsep\fill}lcccccc}
\toprule%
& \multicolumn{2}{@{}c@{}}{FashionStyle14 \cite{TakagiICCVW2017}} & \multicolumn{2}{@{}c@{}}{ShowniqV3} \\\cmidrule{2-3}\cmidrule{4-5}%
Models & Accuracy (\%) & Increment & Accuracy (\%) & Increment \\
\midrule
ResNet50 \cite{resnet} & 71.5 & &57.4 & \\
ResNet50 + IRSN & \textbf{79.1} & +7.6  & \textbf{64.7} & +7.3   \\
\midrule
ConvNeXt-Tiny \cite{convnext} & 74.8 & & 61.2 & \\
ConvNeXt-Tiny + IRSN & \textbf{79.4} & +4.6  & \textbf{65.3} & +4.1   \\
\midrule
ConvNeXt-Base \cite{convnext} & 78.6 & & 64.2 & \\
ConvNeXt-Base + IRSN & \textbf{81.3} & +2.7  & \textbf{66.7} & +2.5        \\
\midrule
EfficientNet-B3 \cite{efficientnet} & 64.3 & & 47.4 & \\
EfficientNet-B3 + IRSN & \textbf{78.8} & +14.5 & \textbf{62.5} & +15.1        \\
\midrule
MobileNetV2 \cite{mobilenetv2} & 72.6 & & 52.6 & \\
MobileNetV2 + IRSN    & \textbf{80.2} & +7.6  & \textbf{61.9} & +9.3        \\
\midrule
Swin-Base \cite{swin} & 77.8 & & 60.2 &      \\
Swin-Base + IRSN       & \textbf{82.0} & +4.2  & \textbf{67.5} & +7.3        \\
\botrule
\end{tabular*}
\end{table}

\begin{figure}[h!]
\centering
\includegraphics[width=0.7\textwidth]{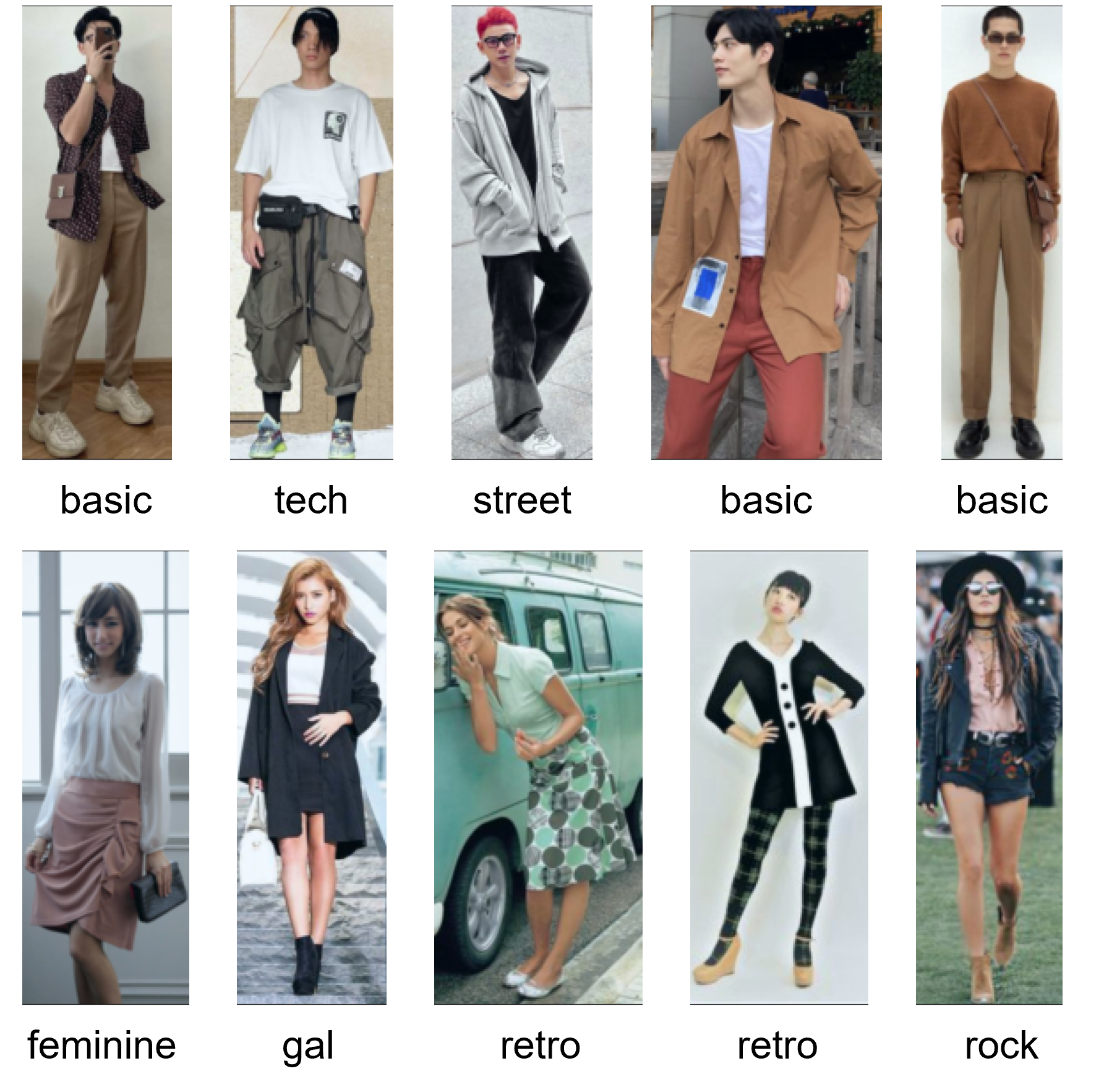}
\caption{Examples of samples correctly classified by IRSN}\label{fig6}
\end{figure}

We compared the performance of IRSN based on six backbones: ResNet \cite{resnet}, ConvNeXt \cite{convnext}, EfficientNet \cite{efficientnet}, MobileNetV2 \cite{mobilenetv2}, and Swin Transformer \cite{swin}, with the corresponding baseline models that only combine backbones and classification heads. Table~\ref{tab2} presents our main results for the FashionStyle14 \cite{TakagiICCVW2017} and ShowniqV3 datasets. IRSN outperformed the baseline models for all six backbones. IRSN exhibited an average of 6.9\% and 7.6\% higher performance than the baseline models for FashionStyle14 and ShowniqV3, respectively. The `EﬀicentNet-B3 + IRSN' model exhibited the largest improvement of 14.5\% on FashionStyle14 and 15.1\% on ShowniqV3 over the baseline. The `Swin-Base + IRSN' exhibited the highest accuracy of 82.0\% on FashionStyle14 and 67.5\% on ShowniqV3. These results are 4.5\% and 7.3\% higher than those of the baseline models. The `ConvNeXt-Base + IRSN' model exhibited the smallest improvement of 2.5-2.7\%. Fig. \ref{fig6} displays a few examples of correctly classified samples. The first row images are correctly classified on ShowniqV3, and the second row images are correctly classified on FashionStyle14 \cite{TakagiICCVW2017}. In particular, the recognition performance for basic class on ShowniqV3 has improved significantly.

\begin{figure}[h!]
\centering
\includegraphics[width=0.9\textwidth]{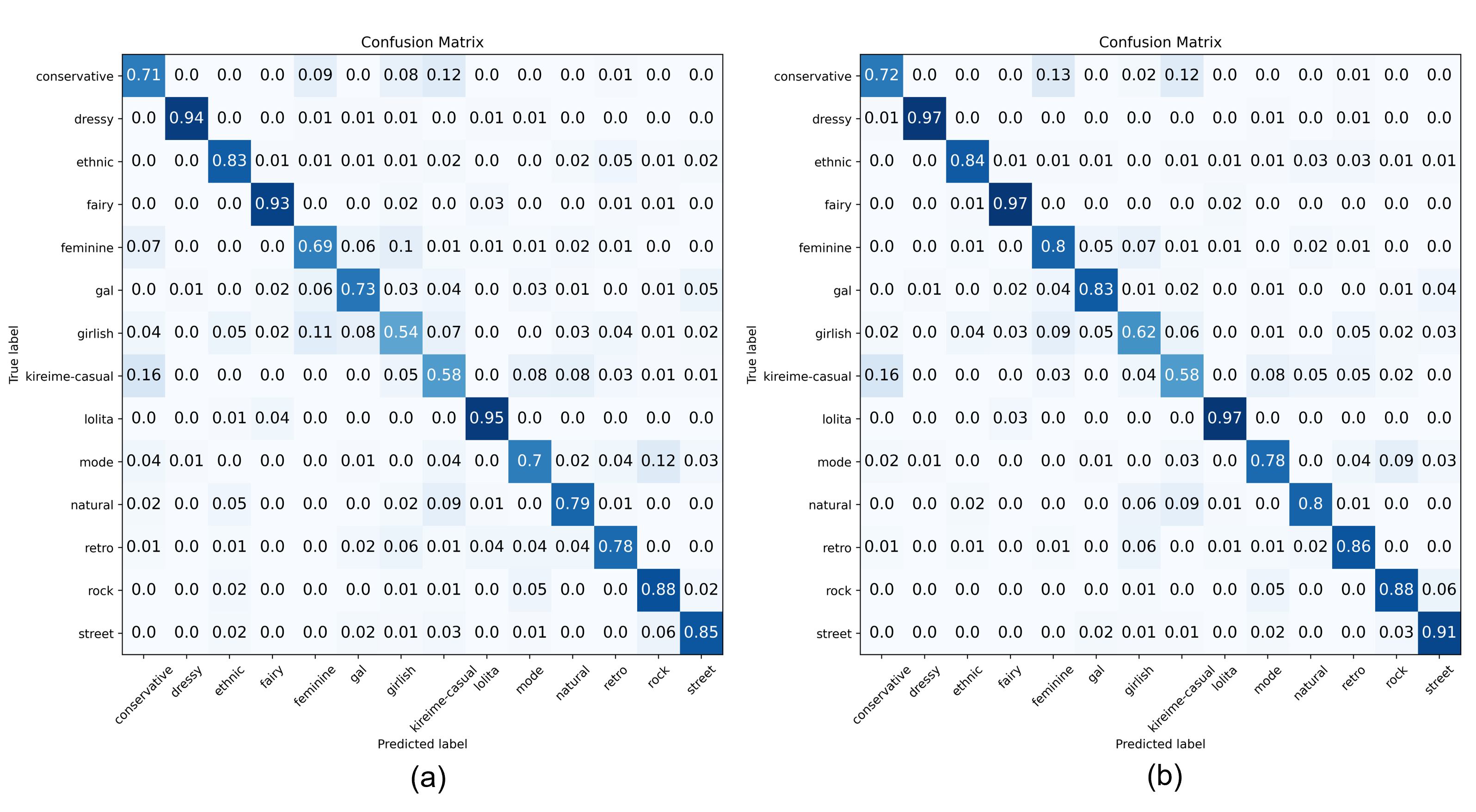}
\caption{Confusion matrix on FashoinStyle14 \cite{TakagiICCVW2017} (a) is the confusion matrix of Swin-Base, (b) is the confusion matrix of Swin-Base with IRSN, respectively.}\label{fig7}
\end{figure}

\begin{figure}[h!]
\centering
\includegraphics[width=0.9\textwidth]{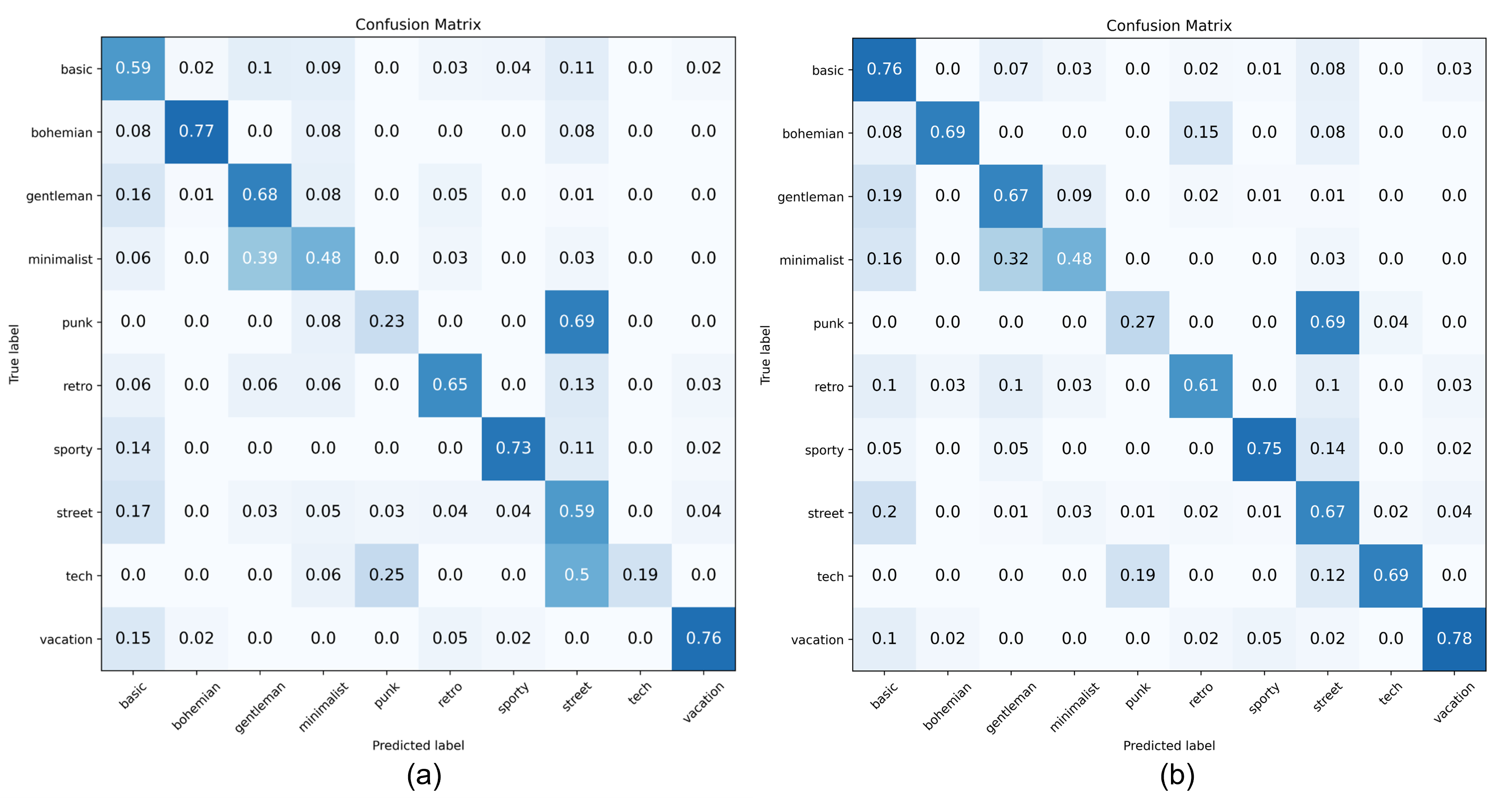}
\caption{Confusion matrix on ShowniqV3 dataset (a) is the confusion matrix of Swin-Base, (b) is the confusion matrix of Swin-Base with IRSN, respectively.}\label{fig8}
\end{figure}

Fig.~\ref{fig7} and Fig.~\ref{fig8} display the confusion matrices of IRSN, which exhibited the largest improvement over their baseline models on the FashionStyle14 \cite{TakagiICCVW2017} and ShowniqV3, respectively. IRSN showed improved accuracy for most of the classes. In the experiment on FashionStyle14 \cite{TakagiICCVW2017}, which consists of women's fashion images, the confusion between conservative and retro classes was significantly reduced. The experiment on the ShowniqV3 of men's fashion images showed the largest reduction in confusion between tech and street style.

\begin{figure}[h]
\centering
\includegraphics[width=0.8\textwidth]{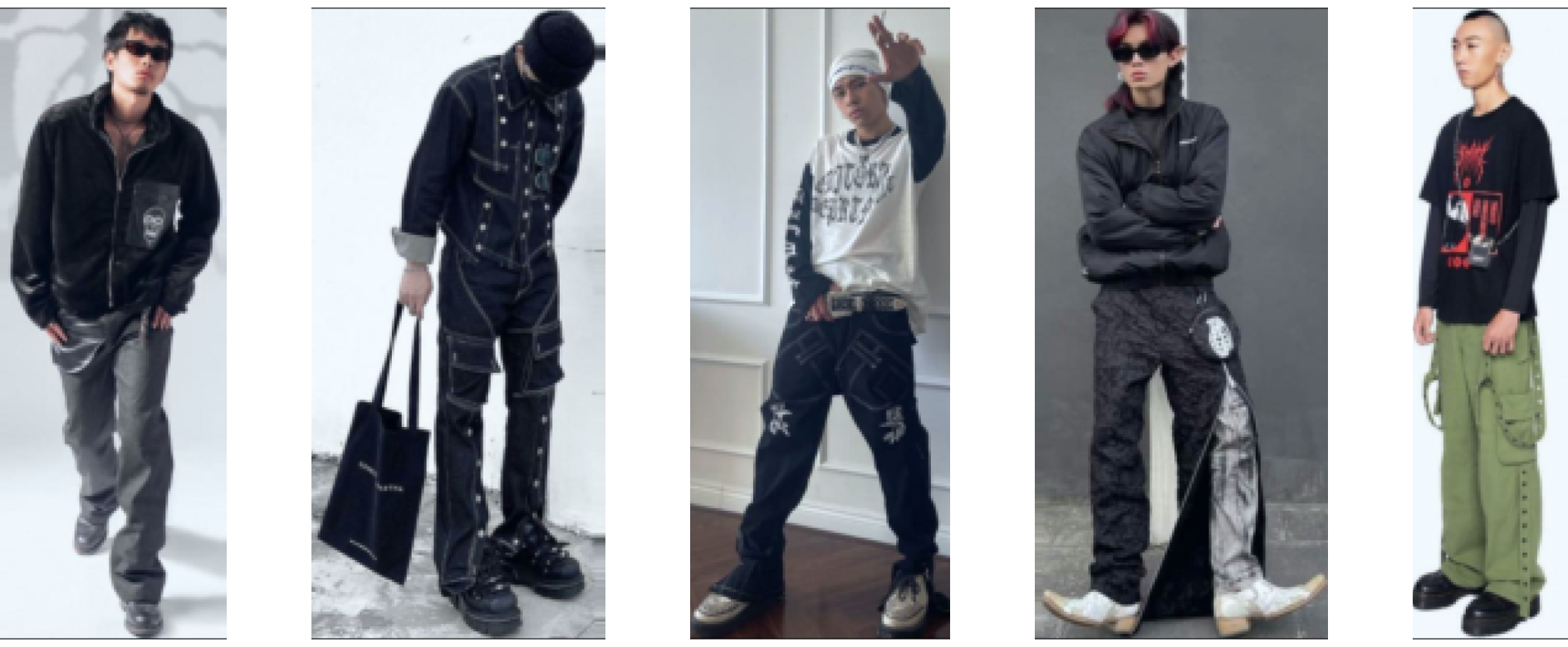}
\caption{Punk style samples misclassified as street style}\label{fig9}

\end{figure}

On the other hand, on ShowniqV3, the confusion from the punk class to the street class did not decreased. Fig.~\ref{fig9} displays a few punk style sample misclassified as the street classes. Although these samples are punk style, they have some street style characteristics, such as a loose fit. The asymmetric error pattern is due to the fact that the street class has significantly more samples than the punk class in the training data.

\begin{table}[]
\caption{Comparison with prior work \cite{dualattention}}\label{tab3}%
\begin{tabular}{@{}lcccccc@{}}
\toprule
Models & FashionStyle14 \cite{TakagiICCVW2017}  & ShowniqV3\\
\midrule
ConvNeXt-Tiny    & 74.8   & 61.2\\
ConvNeXt-Tiny + DualAtt \cite{dualattention}    & 75.4   & 61.6\\
ConvNeXt-Tiny + IRSN    & 79.4   & 65.3\\
\midrule
Our best model & \textbf{82.0} & \textbf{67.5}\\
\botrule
\end{tabular}
\end{table}

We also compared IRSN with previous work on fashion style classification \cite{dualattention}. Few previous studies have been conducted on fashion style classification. In \cite{dualattention}, Chen et al. proposed dual attention that combines criss-cross attention and spatial attention. They applied dual attention to ConvNeXt-Tiny and reported an accuracy of 69.66\% on an in-house dataset, Style10. For comparison, we built `ConvNeXt-Tiny + DualAtt' model following \cite{dualattention} and evaluated it on FashionStyle14 \cite{TakagiICCVW2017} and ShowniqV3. Table~\ref{tab3} displays the results. Applying dual attention to ConvNeXt-Tiny improved the accuracy by 0.6\% and 0.4\% on the two datasets, respectively. IRSN exhibited significantly greater improvement of 4.6\% and 4.1\%.

\subsection{Ablation study}\label{subsec4_4}

To evaluate the effectiveness of the proposed methods, we measured the performance degradation when some modules of IRSN were removed. In the ablation study, we used the `Swin-Base + IRSN' model, which performed the best on the FashionStyle14 \cite{TakagiICCVW2017}, and additionally created three comparison models by removing the AAP, IRP, and the general feature extractor from the `Swin-Base + IRSN'. For the model without AAP, we used GAP instead to reduce the feature dimension. The model without IRP recognizes styles using only the global features extracted by the dual-backbone.

\begin{table}[h]
\caption{The result of the ablation study}\label{tab4}%
\begin{tabular}{@{}lcccccc@{}}
\toprule
Models & Accuracy & Decrement \\
\midrule
Swin-Base + IRSN    & \textbf{82.0}   & \textendash\\
replacing AAP with GAP    & 80.9   & -1.1\\
w/o IRP    & 80.6   & -1.4\\
w/o CLIP vision encoder & 80.6 &  -1.4\\
\botrule
\end{tabular}
\end{table}

The results of the ablation study are presented in Table~\ref{tab4}. Removing IRP or CLIP vision encoder resulted in the largest performance degradation of 1.4\%. 
When the AAP was replaced by GAP, the recognition accuracy decreased by 1.1\%.

\begin{table}[h]
\caption{Accuracy on FashionStyle14 according to the spatial resolution of AAP}\label{tab5}%
\begin{tabular}{@{}lcccccc@{}}
\toprule
Models & Accuracy & Decrement \\
\midrule
Swin-Base + IRSN $\rightarrow$ AAP ($5\times3$)    & \textbf{82.0}   & \textendash\\
AAP ($5\times1$)    & 81.4   & -0.6\\
GAP ($1\times1$)    & 80.9   & -1.1\\
\botrule
\end{tabular}
\end{table}

To further analyze the effect of AAP, we measured performance by varying the spatial resolution of AAP to $5 \times 3$, $5 \times 1$, and $1 \times 1$. Table \ref{tab5} presents the results. When the horizontal resolution was reduced from three to one, the accuracy was decreased by 0.6\%. Reducing both horizontal and vertical resolution decreased the accuracy by 1.1\%. These results suggest that preserving spatial information is crucial in fashion style classification.

\begin{figure}[h!]
\centering
\includegraphics[width=0.8\textwidth]{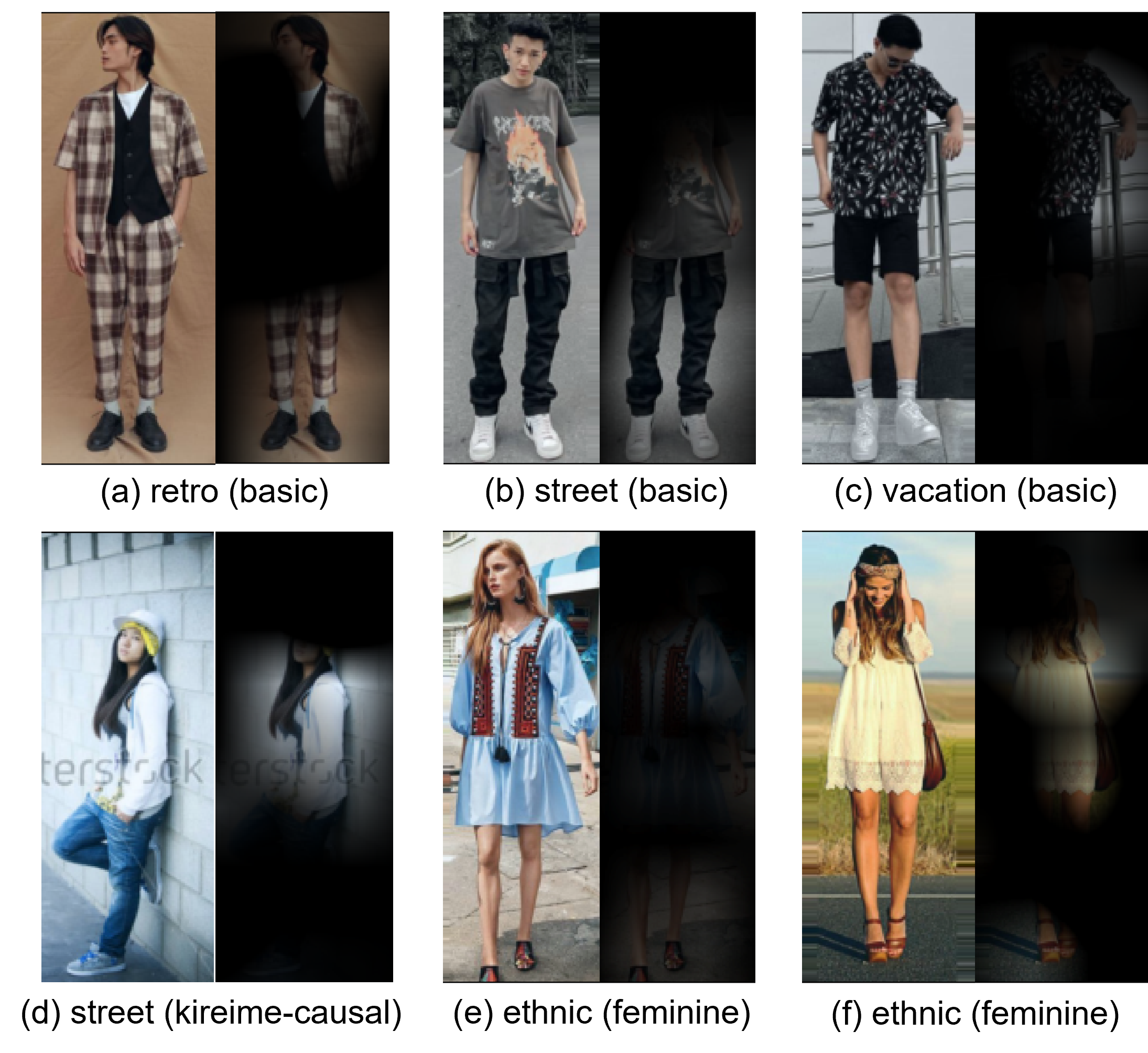}
\caption{
\textbf{The differences between Grad-CAMs extracted using `EfficientNet-B3 + IRSN' and its baseline, EfficientNet-B3.} In (a)-(f), the left images show the samples that were misclassified by EfficientNet-B3 but correctly classified by `EfficientNet-B3 + IRSN.' The right images show the differences between the Grad-CAMs \cite{gradcam} extracted using `EfficientNet-B3 + IRSN' and EfficientNet-B3. Each image was labeled by the correct class. The style names in the parentheses denote the prediction results of EfficientNet-B3.}\label{fig10}
\end{figure}

\subsection{Visualization analysis}\label{subsec4_5}

To understand the behavior of IRSN in more detail, we conducted a visualization analysis using Grad-CAM \cite{gradcam}. We first extracted Grad-CAMs from the samples that were misclassified by EfficientNet-B3 but correctly classified by `EfficientNet-B3 + IRSN' using the two models. Then, we subtracted the Grad-CAM of EfficientNet-B3 from that of `EfficientNet-B3 + IRSN'. Fig. \ref{fig10} displays the difference of Grad-CAMs multiplied by the original images.

In (a), IRSN focused on the check pattern and shoes more than the baseline model to discriminate retro from basic. In (b), IRSN more focused on the loose fit of the trousers than the baseline model to separate street from basic. IRSN focused on the pattern of the top to recognize vacation as in (c) and the hooded zip-up to recognize street in (d), In (e), IRSN considered the top and shoes with folkloric motifs more important in discriminating ethnic from feminine. In (f), IRSN separated ethnic from feminine by focusing on the folklore pattern on the hem of the dress and the leather shoes.

\section{Conclusion}\label{sec5}

In this study, we proposed an item region-based style classification network (IRSN) to effectively classify fashion styles. Similar to how fashion experts classify styles by referring to both the global shape of the fashion image and the characteristics of each item, IRSN separates item features from global feature maps using item region pooling (IRP), processes them with separate item encoders, and then integrates the results by gated feature fusion (GFF). To improve feature extraction performance and robustness, IRSN combines a domain-specific feature extractor trained on fashion images with a general feature extractor trained on a general large-scale image-text dataset.

In experiments, applying IRSN to six widely used backbones improved style classification accuracy by an average of 6.9\% and a maximum of 14.5\% on the FashionStyle14 \cite{TakagiICCVW2017}, and by an average of 7.6\% and a maximum of 15.1\% on the ShowniqV3. The visualization results also confirm that IRSN catches the differences between similar styles better than the baseline model.

\bibliography{sn-bibliography}

\end{document}